\documentclass[11pt]{article}
\usepackage{subcaption}
\usepackage{amsmath,amsfonts,amssymb}
\usepackage{graphicx}
\usepackage{setspace}
\usepackage{csquotes}
\usepackage{hyperref}

\title{A Counterexample in Cross-Correlation Template Matching}

\author{Serap A.~Savari \\ Texas A\&M University, College Station, TX 77843-3128, USA}

\begin{document}

\maketitle

\begin{abstract}
  Sampling and quantization are standard practices in signal and image
  processing, but a theoretical understanding of their impact is incomplete.
  We consider discrete image registration when the underlying function is a
  one-dimensional spatially-limited piecewise constant function.
  For ideal noiseless sampling the number of samples from each region of
  the support of the function generally depends on the placement of the
  sampling grid.  Therefore, if the samples of the function are noisy,
  then image registration requires alignment and segmentation of the
  data sequences.  One popular strategy for aligning images is selecting
  the maximum from cross-correlation template matching.  To motivate more
  robust and accurate approaches which also address segmentation, we provide
  an example of a one-dimensional spatially-limited piecewise constant
  function for which the cross-correlation technique can perform poorly on
  noisy samples.  While earlier approaches to improve the method involve
  normalization, our example suggests a novel strategy in our setting.
  Difference sequences, thresholding, and dynamic programming are well-known
  techniques in image processing.  We prove that they are tools to correctly
  align and segment noisy data sequences under some conditions on the noise.
  We also address some of the potential difficulties that could arise in a
  more general case.
\end{abstract}

\section{Introduction}
  The alignment of images of the same scene is needed for applications
  as diverse as medical imaging, hyperspectral image processing, and
  micro/nano analysis \cite{tong}.
  Image registration is an extensively studied topic which uses
   models of image similarity or transformation to determine how to match
images, and it is often influenced by computational considerations
 \cite{tong}, \cite[\S 2.5]{ip}.  In \cite{savari}, \cite{reference} we offer
 a different perspective. Digital images are sampled \cite[\S 2.4]{ip},
 and Nyquist's sampling theorem \cite{nyquist} states that the possibility of
 reconstructing a signal from samples is constrained by properties of that
 signal.  Therefore, properties of a signal may also limit how well we can
 correlate information about it from different sets of samples for image
 registration tasks.  Digital images are quantized \cite[\S 2.4]{ip}, and we
 investigate how quantization impacts digital image registration in the
 simplest setting.  
  Discrete image registration seeks to correlate information based on samples.
  In \cite{savari} we mainly   concentrate on the static forward problem
  of describing the ideal sampling patterns 
 of the support of a spatially limited piecewise constant one-dimensional
 function in the absence of noise; in \cite{savari} we also show that
 the inverse problem of determining subinterval relationships from
 sampling patterns in the absence of additional assumptions is related
 to partitions of a   unit hypercube.  In \cite{reference}, we
 extend the results of \cite{savari} to generate an estimate of the
 underlying function from ideal and noiseless sets of sampling patterns
 in the absence of additional assumptions.
 We mainly focus there on minimizing the energy of the estimation error.
 The unusual outcome of \cite{reference} is that the 
  uncertainties associated with the locations of the discontinuity points
  of the function depend on the reference  point of the function; hence,
 the accuracy of the estimate generally
  depends on the reference point.

  In this work, we begin to extend the foundation of \cite{savari}
  and \cite{reference} to the case where the samples are corrupted by noise.
  We focus on the case of two sets of noisy observations, but in \cite{savari}
  and \cite{reference} we consider an arbitrary number of noiseless
  observations.  It is well known that the data sequences will generally
  need alignment.  In \cite{savari} we discuss how the number of
  samples from each region of the support of the function is based on
  the position of the sampling grid.  Therefore data sequences with noisy
  samples will also need segmentation of the samples into regions.
  There are multiple approaches to the alignment of images \cite{tong},
  \cite[p. 1061]{ip}, \cite[Appendix B]{stem}.
  The most basic of these is selecting the maximum from cross-correlation
  template matching, which in the case of one-dimensional signals involves
  taking inner products of components of a data sequence with shifted
  versions of another.  The alternatives described in \cite{tong},
  \cite[p. 1061]{ip}, \cite[Appendix B]{stem} involve various types of
  normalizations, and \cite[Appendix B]{stem} recommends the use of
  standard cross-correlations in a microscopy application for very noisy
  data.  We focus here on the pure cross-correlation technique to motivate
  the ones that we propose.  We show a spatially limited piecewise constant
  function and two data sequences each consisting of nine noisy samples
  of the function for which the cross-correlation technique performs
  poorly. Cauchy's inequality (see, e.g., \cite{hardy}) is a well-known
  upper bound on the inner product of two sequences of real numbers, and
  it motivates finding representations of general noisy data sequences
  that have transformations into more similar sequences.
  We apply the well-known techniques of difference sequences, thresholding
  \cite[\S 10.3]{ip} and sometimes dynamic programming \cite{segmentation}
  as the basis for such transformations to handle both alignment and
  segmentation.  In some cases, we can prove that the proposed techniques
  offer optimal alignment and segmentation. In other situations, we describe
  possible difficulties and approaches to address some of them.
  Under the assumption of zero-mean additive noise which is independent
  from sample to sample we can apply the results of \cite{reference} to
  the aligned and segmented samples to estimate the underlying function.

  The plan for the rest of the paper is as follows.  In Section~2, we provide
  our motivating example and show a class of additive noise for which
  cross-corrleation template matching does not result in a suitable
  alignment.  In Section~3, we review
  some of the notation and results from \cite{savari} and \cite{reference}.
  In Section~4, we discuss difference sequences.  In Section~5, we propose
  a scheme using thresholding on difference sequences for alignment
  and segmentation.
  In Section~6, we consider a dynamic programming
  framework for difference sequences for alignment
  and segmentation in more general settings.
  In Section~7, we conclude.

  \section{A Small Example}
  Consider the underlying spatially-limited piecewise constant function
  \begin{displaymath}
g(t) \; = \; \left\{ \begin{array}{ll}
1, & 0 \leq t < 1.3T, \; 2.75T \leq t < 4.1T \\
-1, & 1.3T \leq t < 2.75T, \; 4.1T \leq t < 5.4T \\
0, & \mbox{otherwise.}
\end{array}
\right.
  \end{displaymath}
  Suppose the sampling interval is $T$, the first sequence $\gamma_1$
  of nine noiseless samples of $g(t)$ has its first sample at $t=-0.95T$,
  and the second sequence $\gamma_2$
  of nine noiseless samples of $g(t)$ has its first sample at $t=-0.5T$.
  Then $\gamma_1$ and $\gamma_2$ are
  \begin{eqnarray*}
    \gamma_1 & = & (\gamma_1 [1], \dots ,  \gamma_1 [9]) \; = \;
    (0, \ 1, \ 1, \ -1, \ 1, \ 1, \ -1, \ 0, \ 0) \\
    \gamma_2 & = & (\gamma_2 [1], \dots ,  \gamma_2 [9]) \; = \;
    (0, \ 1, \ -1, \ -1, \ 1, \ -1, \ 0, \ 0, \ 0).
      \end{eqnarray*}

  There is a collection of piecewise constant functions and sampling grid
  positions that would result in $\gamma_1$ and $\gamma_2$, so without
  prior knowledge about $g(t)$ and/or the translations of the sampling grid
  from one sequence of samples to the next there are limits to the accuracy
  of estimates pertaining to $g(t)$ that can be obtained by the components
  of $\gamma_1$ and $\gamma_2$ corresponding to the support of $g(t)$,
  and we study this more generally in \cite{savari} and \cite{reference}.
  Note that in this example the samples corresponding to the beginning of the
  support of $g(t)$ are aligned, but this need not be the case in general.

  Let $e_i  = ( e_i [1], \ \dots , \ e_i [9] ), \; i \in \{1, \ 2\}$ denote
  the additive noise corrupting $\gamma_i$.
  Suppose the underlying noise model for some $x \in [0, \ 0.5]$ is
  \begin{displaymath}
    P[e_i [n] = x] \; = \; P[e_i [n] = -x] \; = \; 0.5, \; i \in \{ 1, \ 2\},
    n \in \{ 1, \ 2, \ \dots , \ 9 \}.
  \end{displaymath}

  We consider the following error vectors:
    \begin{eqnarray*}
      e_1 & = &     (-x, \ x, \ -x, \ x, \ x, \ -x, \ -x, \ x, \ x) \\
      e_2 & = &     (-x, \ -x, \ x, \ -x, \ -x, \ x, \ x, \ x, \ -x) .
    \end{eqnarray*}

    Our observed sequences of corrupted samples
    $y_i  = ( y_i [1], \ \dots , \ y_i [9] ), \; i \in \{1, \ 2\}$ satisfy
    \begin{displaymath}
      y_i [n] \; = \; \gamma_i [n] + e_i [n] ,  \; i \in \{ 1, \ 2\},
    n \in \{ 1, \ 2, \ \dots , \ 9 \}
    \end{displaymath}
    so that
    \begin{eqnarray*}
      y_1 & = &     (-x, \ 1+x, \ 1-x, \ -1+x, \ 1+x, \ 1-x, \ -1-x, \ x, \ x) \\
      y_2 & = &     (-x, \ 1-x, \ -1+x, \ -1-x, \ 1-x, \ -1+x, \ x, \ x, \ -x);
    \end{eqnarray*}
    assume $y_1 [n] = y_2 [n] = 0, \; n \not\in \{ 1, \ 2, \ \dots , \ 9 \}$.
    To consider cross-correlation template matching, let
    \begin{displaymath}
      r_{y,12} [i] \; = \; \sum_n y_1 [n] y_2 [n+i] , \; i \in
      \{-8, \ -7, \ \dots , \ 7, \ 8\}.
    \end{displaymath}

    Table 1 displays $r_{y,12} [i]$ as a function of $x$.

    \begin{table}
    \begin{center}
\begin{tabular}{|l|l|l|}
\hline
-8: $-x^2$ & -2: $-4+5x^2$ & 4: $-1-2x+x^2$ \tabularnewline
\hline
-7: $x-2x^2$ & -1: $3-9x+2x^2$ & 5: $ 4x-2x^2$ \tabularnewline
\hline
-6: $x+x^2$ & 0: $ 1+3x-5x^2$ & 6: $ x^2$ \tabularnewline
\hline
-5: $-1-3x+2x^2$ & 1: $ -4+x$ & 7: $ -x-2x^2$ \tabularnewline
\hline
-4: $2-3x-3x^2$ & 2: $1-2x+x^2$ & 8: $x^2$ \tabularnewline
\hline
-3: $1+5x-2x^2$ & 3: $2x+2x^2$ & \mbox{} \tabularnewline
\hline
\end{tabular}
\par\end{center}
\caption{Cross-correlations as a function of $x$ displayed as $i: r_{y,12} [i]$.}
    \end{table}

    For $0 \leq x \leq 0.5,$
    \begin{displaymath}
      \arg \max_i  r_{y,12} [i]
      \; = \; \left\{ \begin{array}{ll}
        -1, & 0 \leq x \leq \frac{7-\sqrt{41}}{4}
        \\
-3, & \frac{7-\sqrt{41}}{4} \leq x \leq 0.5.
\end{array}
\right.
\end{displaymath}

    At $x=0$, cross-correlation template matching recommends that we shift
    $\gamma_2$ by one to the right relative to $\gamma_1$, which is an error
    for left-to-right processing.  At $x=0.15$, the cross-correlation
    technique recommends that we shift
    $\gamma_2$ by three to the right relative to $\gamma_1$, resulting in
    the loss of a considerable portion of the support of $g(t)$.
    Furthermore, from $\gamma_1$ and $\gamma_2$ it follows that $y_1$ and $y_2$
    need segmentation.  Therefore, our objective here is to consider
    alternatives.

\section{Notation and Some Earlier Results}
In general, our underlying spatially limited piecewise constant function
with $m$ regions in its support is given by \cite{savari} 
\begin{equation}
g(t) \; = \; \left\{ \begin{array}{ll}
g_1, & 0 \leq t < R_1 \\
g_i, & \sum_{j=1}^{i-1} R_j  \leq t < \sum_{j=1}^{i} R_j, \; i \in \{2, \ \dots ,
\ m\} \\
0, & \mbox{otherwise};
\end{array}
\right.
\end{equation}
here $g_1 \neq 0, \; g_m \neq 0,$ and $g_i \neq g_{i+1}, i \in \{1, \ \dots ,
\ m-1\}$,  and we define $g_0 = g_{m+1} = 0 $.
For the sampling interval $T$, region length $R_i$ is given by
\begin{displaymath}
R_i = (n_i-f_i)T, \; i \in \{1, \ \dots , \ m\}, 
\end{displaymath}
where for each $i, \; n_i$ is an integer that is at least two and
$0 < f_i \leq 1$; i.e., we take at least one sample from each region.

Let $\Delta_i \in [0, \ T)$ represent the distance from the left endpoint
 of region $i$ to the first sampling point within the region.
 The number $\eta_i$ of samples from region $i$ depends on $\Delta_i$.
Define $\lfloor x \rfloor , \; x \geq 0,$ as the
integer part of $x$.  For non-negative integers $K$, define
\begin{eqnarray*}
  \kappa (i, \ K) & = & \lfloor \sum_{j=0}^{K} f_{i+j}  \rfloor \\
  d_{i,K} & = &  \sum_{j=0}^{K} n_{i+j} -  \kappa (i, \ K) .
\end{eqnarray*}

The following result is \cite[Proposition 1]{reference}:

\noindent {\bf Proposition 1:} For $i \geq 1$ and $i+K \leq m$, the
cumulative number of samples over regions $i, \ i+1, \ \dots , \ i+K$ satisfies
\begin{displaymath}
\sum_{j=0}^K \eta_{i+j} \; = \; \left\{ \begin{array}{ll}
  d_{i,K}, & 0 \leq \Delta_i < (1+\kappa (i, \ K) - \sum_{j=0}^{K} f_{i+j})T \\
  d_{i,K}-1, &  (1+\kappa (i, \ K) - \sum_{j=0}^{K} f_{i+j})T \leq \Delta_i 
 < T .
\end{array}
\right.
\end{displaymath}

Assume that there are no integers $i$ and $K$ for which 
$\sum_{j=0}^{K} f_{i+j}$ is an integer.  Then it follows from
Proposition~1 that we know $\sum_{j=0}^{K} R_{i+j}$ to within $T$
if our data provides two values of $\sum_{j=0}^{K} \eta_{i+j}$ so that we
know $d_{i,K}$.  Otherwise, we do not know $d_{i,K}$ and we can only infer
$\sum_{j=0}^{K} R_{i+j}$ to within $2T$.

In \cite{reference} we emphasize that the reference point $t=0$ in (1)
is one of $m+1$ discontinuity points of $g(t)$, and for the purpose of
estimation it is better to study translations
$g^{(l)} (t), \ l \in \{0, \ \dots , \ m\}$, of $g(t)$ where the reference
point $t=0$ corresponds to the $l^{\mbox{th}}$ discontinuity point.
Let $D_i^{(l)}, \ l \in \{0, \ \dots , \ m\}$, be the location of discontinuity
point $i$ for this translation.   Let $R_0=0$.  Then 
    \begin{eqnarray*}
      g^{(l)} (t) & = & g(t + \sum_{i=0}^{l} R_i ) \\
      D_i^{(l)} & = &  \left\{ \begin{array}{ll}
        - \sum_{j=i+1}^{l} R_j ,        &  i \in \{0, \ \dots , \ l-1\}     \\
        0, & i=l \\
         \sum_{j=l+1}^{i} R_j ,        &  i \in \{l+1, \ \dots , \ m\}     

\end{array}
\right.
      \end{eqnarray*}

    The main result of \cite{reference} when there are two sequences of samples
    is \cite[Theorem 7]{reference}.  For simplicity we focus here on a special
    case of that result which generalizes \cite[Theorem 5]{reference}.
    Let $U \subseteq \{0, \ \dots , \ m \} \setminus \{l \}$
denote the set of indices for which the location of
  ${D_i^{(l)}}$ is not known to
  within $T$ when $i \in U$, and let $U^c = 
  \{0, \ \dots , \ m \} \setminus U \cup \{l \}$.  $U$ and $U^c$ depend on $l$,
  but we follow the simpler notation of \cite{reference}.
  When $i=l$, let $C_l^{(l)}=D_l^{(l)}=  0$.
  Define integers $C_i^{(l)}, \; i \in U^c$, by
    \begin{equation}
\left\{ \begin{array}{ll}
  \sum_{j=i+1}^{l} \eta_j  \in \{C_i^{(l)}-1, \ C_i^{(l)}\},
  &  i<l, \ i \in U^c 
  \\
  \sum_{j=l+1}^{i} \eta_j \in \{C_i^{(l)}-1, \ C_i^{(l)}\},
  &  i>l, \ i \in U^c 
\end{array}
\right.
    \end{equation}
    for both sets of  sampling patterns.  Then \cite[Equation (4)]{reference}
is
      \begin{displaymath}
           - C_i^{(l)} T \; < \; D_i^{(l)} \;       < \; -(C_i^{(l)}-1)T, \;
     i<l,  \ i \in U^c        
    \end{displaymath}
    \begin{equation}
(C_i^{(l)}-1)T \; < \; D_i^{(l)} \; < \;  C_i^{(l)} T, \;
     i>l, \  i \in U^c.
    \end{equation}
Define integers $C_i^{(l)}, \; i \in U$, by
    \begin{equation}
\left\{ \begin{array}{ll}
  \sum_{j=i+1}^{l} \eta_j  =  C_i^{(l)},   &  i<l, \ i \in U 
  \\
  \sum_{j=l+1}^{i} \eta_j =  C_i^{(l)},   &  i>l, \ i \in U 
\end{array}
\right.
    \end{equation}
    for both sets of  sampling patterns.  Then the counterpart to (3) 
is
      \begin{displaymath}
           - (C_i^{(l)}+1) T \; < \; D_i^{(l)} \;  < \; -(C_i^{(l)}-1)T, \;
     i<l, \   i \in U
    \end{displaymath}
    \begin{equation}
(C_i^{(l)}-1)T \; < \; D_i^{(l)} \; < \;  (C_i^{(l)}+1) T, \;
     i>l, \  i \in U.
    \end{equation}

    To unify the discussion of the previous cases, let $G_{l,R}^{(l)}=0,$ 
and for $i \neq l$ let  $G_{i,L}^{(l)}$     and  $G_{i,R}^{(l)}$ satisfy
    \begin{equation}
      D_i^{(l)} \in (G_{i,L}^{(l)} , \ G_{i,R}^{(l)} ),
          \end{equation}
    where $G_{i,L}^{(l)}$     and  $G_{i,R}^{(l)}$ are specified by (3) when
    $i \in U^c$ and by (5) when $i \in U$.  We will apply the following
    special case of \cite[Theorem 7]{reference}:
    
    \noindent {\bf Theorem 2:} Suppose that whenever $\eta_i =1$ for both
    sampling patterns then $D_{i-1}^{(l)}$ is known to within $T$ if $i>l$
    and $D_{i+1}^{(l)}$ is known to within $T$ if $i<l$.
Then the estimate
\begin{displaymath}
  \hat{g}^{(l)}(t)
  \; = \; \left\{ \begin{array}{ll}
      0, & t \leq G_{0,L}^{(l)} \; \mbox{or} \; t \geq G_{m,R}^{(l)} \\
g_i, & i \neq l, \; G_{i-1,R}^{(l)}  \leq t \leq G_{i,L}^{(l)} \\
g_l, & G_{l-1,R}^{(l)}  \leq t < 0 \\
\frac{g_i + g_{i+1}}{2}, & i \neq l, \;
G_{i,L}^{(l)}  < t < G_{i,R}^{(l)} 
\end{array}
\right.
\end{displaymath}
is an extremum of the problem of minimizing the energy of the estimation
error with respect to $  g^{(l)}(t)$, and the corresponding estimation error
is
\begin{displaymath}
  \sum_{i \in U^c} \left( \frac{g_i - g_{i+1}}{2} \right)^2 T
  +  \sum_{i \in U} \left( \frac{g_i - g_{i+1}}{2} \right)^2 (2T).
  \end{displaymath}
    
\section{Difference Sequences}
Returning to our motivating example, cross-correlation template matching
does not account for the variations within the possible patterns of sampling
counts that can be obtained for the support of $g(t)$.
The Cauchy inequality (see, e.g., \cite{hardy}) states that for sequences
of real numbers $\{ a_i \}$  and $\{ b_i \}$,
\begin{equation}
  | \sum_i a_i b_i | \; \leq \; \sqrt{ \sum_i {a_i}^2 \sum_i {b_i}^2 }
\end{equation}
with equality if and only if there is a constant $k$ for which
$a_i = k b_i$ for all $i$.  In general, $\sum_n (\gamma_1 [n])^2 \neq
\sum_n (\gamma_2 [n])^2$.

Since $g(t)$ is piecewise constant, define the difference sequence
$\delta_i = ( \delta_i [1], \ \dots , \ \delta_i [9])$ of $\gamma_i , \
i \in \{1, \ 2\}$, by
\begin{displaymath}
  \delta_i [n] \; = \; \gamma_i [n] - \gamma_i [n-1], \;
  n \in \{1, \ 2, \ \dots , \ 9\},
  \end{displaymath}
where $\gamma_i [0]= 0, i \in \{1, \ 2\}$.  For our example,
\begin{eqnarray*}
  \delta_1 & = & (0, \ 1, \ 0 , \ -2 , \ 2, \ 0 , \ -2, \ 1, \ 0) \\
  \delta_2 & = & (0, \ 1, \ -2 , \ 0, \ 2,  \ -2, \ 1, \ 0, \ 0)
\end{eqnarray*}
$\delta_1$ and $\delta_2$ have the same subsequence of non-zero values,
which correspond to the first samples after discontinuity points of $g(t)$.
The zero values indicate the continuation of a constant region.
More generally, we have

\noindent {\bf Proposition 3:} Let
$\gamma_i = (\gamma_i [1], \ \dots , \ \gamma_i [N]), \; i \in \{1, \ 2\}$,
be uncorrupted data sequences containing samples of the support of $g(t)$
and possibly additional samples. Let
$\delta_i = ( \delta_i [1], \ \dots , \ \delta_i [N]), \
i \in \{1, \ 2\}$, be defined by
$  \delta_i [n] \; = \; \gamma_i [n] - \gamma_i [n-1], \;
n \in \{1, \ 2, \ \dots , \ N\},$
where $\gamma_i [0]= 0, \ i \in \{1, \ 2\}$.
Then the subsequence of non-zero components of $\delta_1$ and $\delta_2$
are each $\tau_1 , \ \dots , \ \tau_{m+1}$, where
$\tau_k = g_k - g_{k-1}$ for all $k$.  The position of $\tau_k$ within
$\delta_i , \ i \in \{1, \ 2\},$ indicates the first sample of value $g_k$
following the $k^{\mbox{th}}$ discontinuity point of $g(t)$.

\noindent {\bf Proof:}  Let $l_{i,j}, \ i \in \{1, \ 2\}, \
j \in \{1, \ \dots , \ m+1\}$, denote the location in $\gamma_i$ of the
first sample of value $g_j$ following the
$j^{\mbox{th}}$ discontinuity point of $g(t)$.
Then for $i \in \{1, \ 2\}, \ \lambda \in \{1, \ \dots , \ N\}$, 
    \begin{equation}
\gamma_i [\lambda ] = \left\{ \begin{array}{ll}
  0,        & \lambda < l_{i,1} \; \mbox{or} \; \lambda \geq l_{i,m+1}   \\
  g_k ,     & l_{i,k} \leq \lambda < l_{i,k+1}, \; 
 k \in \{1, \ \dots , \ m\}  .
\end{array}
\right.
    \end{equation}
    Therefore,
        \begin{equation}
\delta_i [\lambda ] = \left\{ \begin{array}{ll}
  g_k - g_{k-1},        & \lambda = l_{i,k}, \;  k \in \{1, \ \dots , \ m+1\}  
  \\
  0 ,     & \mbox{otherwise,}
\end{array}
\right.
    \end{equation}
and the subsequence of non-zero components of $\delta_i$ is
\begin{displaymath}
  g_1, \ g_2-g_1, \ \dots g_m-g_{m-1}, \ - g_m .
  \end{displaymath}
Finally, by (8) and (9) the positions $l_{i,k}$ within $\delta_i$ determine
the first sample of value $g_k$ in $\gamma_i$
following the $k^{\mbox{th}}$ discontinuity point of $g(t). \; \Box$

To extend difference sequences to noisy data sequences, define
$d_i = ( d_i [1], \ \dots , \ d_i [N]), \ i \in \{1, \ 2 \}$, by
$d_i [n] \; = \; y_i [n] - y_i [n-1], \; n \in \{1, \ \dots , \ N \}$.
For our example,
\begin{eqnarray*}
  d_1 & = & (-x, \ 1+2x, \ -2x , \ -2+2x , \ 2, \ -2x , \ -2, \ 1+2x, \ 0) \\
  d_2 & = & (-x, \ 1, \ -2+2x , \ -2x, \ 2,  \ -2+2x, \ 1, \ 0, \ -2x)
\end{eqnarray*}
It is possible to show that for $0.15 \leq x \leq 0.5$, the cross-correlation
template matching technique on $d_1$ and $d_2$ would again recommend shifting
$d_2$ by three to the right relative to $d_1$.

\section{A Thresholding Scheme}
We have the following result,

\noindent {\bf Theorem 4:} Suppose we have two difference sequences
$d_1$ and $d_2$ with respective ground truth difference sequences
$\delta_1$ and $\delta_2$.  Define
\begin{displaymath}
  \tau \; = \; \max_{0 \leq k \leq m} |g_{k+1}-g_k |.
\end{displaymath}
Let $Z_i \subset \{ 1, \ \dots, \ N \}, \ i \in \{1, \ 2\}$, be the set
of indices for which $\delta_i [j]  = 0$ if $j \in Z_i$ and
$\delta_i [j]  \neq 0$ if $j \not\in Z_i$.
Suppose there is $v \in (0, \ \tau )$ for which
\begin{itemize}
\item $|d_i [j]| < v, \ i \in \{1, \ 2 \}$, if $j \in Z_i$,
\item $d_i [j] \geq v, \ i \in \{1, \ 2 \}$, if $j \not\in Z_i$
  and $\delta_i [j] > 0$,
  \item $d_i [j] \leq -v, \ i \in \{1, \ 2 \}$, if $j \not\in Z_i$
  and $\delta_i [j] <  0$,
\end{itemize}
Then a thresholding scheme that sets all components of $d_i, \ i \in \{1, \ 2\}$,
with magnitude less than $v$ to zero results in a sequence that identifies
the location of the $m+1$ samples within $y_i$ and $\gamma_i$ which
immediately follow the discontinuity points of $g(t)$.

\noindent {\bf Proof:} The scheme will by definition identify the exact set
of zero components in $\delta_1$ and $\delta_2$ and the signs of the non-zero
components of $\delta_1$ and $\delta_2$.  By Proposition~3, the locations
of the non-zero components of $\delta_1$ and $\delta_2$ indicate where the
$m+1$ regions of interest of $g(t)$ begin within $\gamma_1$ and $\gamma_2 .
\; \Box$

For our exmaple, Theorem~4 applies when $0 \leq x < 0.5$ and $v=1$.
If a thresholding scheme is possible in general, then the threshold must
satisfy certain constraints.  A pair of thresholded sequnces must have the
same number of non-zero terms and the same sign for the $i^{\mbox{th}}$
non-zero terms for every $i$.  Furthermore, the resulting inferences on
patterns of sample counts must be consistent with Proposition~1 and the
results of \cite{savari}.  One could enumerate the components of $d_1$
and $d_2$ from smallest to largest magnitudes to progressively increase
the value of the threshold until there is either a candidate threshold
that satisfies the constraints or there is a determination that the
thresholding scheme is infeasible.

For our example with $0 \leq x < 0.5$ and $v=1$, the thresholded sequences
are
\begin{eqnarray*}
  &  & (0, \ 1+2x, \ 0 , \ -2+2x , \ 2, \ 0 , \ -2, \ 1+2x, \ 0) \\
  &  & (0, \ 1, \ -2+2x , \ 0, \ 2,  \ -2+2x, \ 1, \ 0, \ 0) .
\end{eqnarray*}
The number of non-zero terms in each imply that $m=4$, so we must estimate
$(g_1 , \ g_2, \ g_3 , \ g_4)$.
By studying the positions of the non-zero terms, we can correctly identify
which components of each sequence correspond to each region of the support
of $g(t)$, and we average the corresponding terms of $y_1$ and $y_2$ to
obtain the estimate
\begin{displaymath}
  (\hat{g}_1 , \ \hat{g}_2, \ \hat{g}_3 , \ \hat{g}_4)   \; = \;
  \left( 1 - \frac{x}{3}, \; -1 + \frac{x}{3}, \; 1 - \frac{x}{3}, \; -1 \right) .
\end{displaymath}

Theorem 2 suggests that we select $l=2$ and
        \begin{displaymath}
\delta_i [\lambda ] = \left\{ \begin{array}{ll}
  0.5 - \frac{x}{6},        & -4T < t < -2T   \\
  1 - \frac{x}{3},        & t=-2T, \; 0 < t \leq T   \\
  -1 + \frac{x}{3},        & -T \leq  t < 0   \\
  - \frac{x}{6},        & T < t < 2T   \\
  -1, & t=2T \\
  -0.5 ,        & 2T < t < 3T   \\
  0 ,     & \mbox{otherwise,}
\end{array}
\right.
        \end{displaymath}
        The energy of the difference of $\hat{g}^{(2)} (t)$ at $x=0$
        and the limit of $\hat{g}^{(2)} (t)$ when $x$ approaches $0.5$
        from below is $\frac{11T}{144}$.

        At $x=0.5$, the thresholding scheme no longer works; here,
  \begin{eqnarray*}
  d_1 & = & (-0.5, \ 2, \ -1 , \ -1 , \ 2, \ -1 , \ -2, \ 2, \ 0) \\
  d_2 & = & (-0.5, \ 1, \ -1 , \ -1, \ 2,  \ -1, \ 1, \ 0, \ -1).
  \end{eqnarray*}
  If $0.5 < v \leq 1$, then the thresholded sequences each have seven non-zero
  terms, but the signs differ for the sixth and seventh pairs of terms.
  If $1 < v < 2$, then the first thresholded sequence has four non-zero terms
  and the second has one.  Therefore, we need a different technique to
  process $d_1$ and $d_2$ in this case.

  \section{A Dynamic Programming Scheme}
  Dynamic programming is a well-known tool for image segmentation
  (see, e.g., \cite{segmentation}) and for the decoding of corrupted data
  in digital communication \cite{viterbi}.
  We formulate a longest path problem from a starting vertex to a termination
  vertex in a directed, acyclic graph to seek the most desirable equal length
  subsequences of $d_1$ and $d_2$ subject to some of the constraints on
  patterns of sampling counts; additional constraints can be included at the
  cost of increased computational complexity.  The graph also contains
  $2N-1$ alignment vertices which are neighbors of the starting vertex and
  $3N^2-4N+2$ vertices for potential segmentation pairs within $d_1$
  and $d_2$ which include some information about past segmentations.
  The alignment vertices are labeled by ordered pairs
  \begin{displaymath}
    (j_1 , \ j_2) \in \{ (0, \ 0), \ (0, \ 1), \ \dots , \ (0, \ N-1), \
    (1, \ 0), \ \dots , (N-1, \ 0) \},
  \end{displaymath}
  where vertex $(j_1 , \ j_2)$ signifies that we begin examining $d_i, \
  i \in \{1, \ 2\}$, in position $j_i$ and position $0$ precedes the
  sequence.  We propose three edge weight functions; for all of them any
  edge from the starting vertex to an alignment vertex has weight zero.
  Each alignment vertex is also neighbors with a subset of the segmentation
  vertices.

  The segmentation vertices and edges associated with them are motivated by
  two properties from Proposition~1.  First, the number $\eta_i, \
  i \in \{1, \ \dots , \ m \}$, of samples from region $i$ of the support of
  $g(t)$ can differ in two sets of samples by at most one, and the total
  number of samples from the first $i$ regions of the support of $g(t)$ can
  also differ by at most one.  A segmentation pair $(d_1 (j_1 ), \
  d_2 (j_2 )), \; j_1, \ j_2 \in \{2, \ \dots , \ N \}$, marking the
  beginning of the same region in the support of $g(t)$ or in the final
  all-zero region is associated with three vertices in the graph labeled
  $(j_1, \ j_2, \ 0), \; (j_1, \ j_2, \ 1),$ and $(j_1, \ j_2, \ 2),$
  The vertex $(j_1, \ j_2, \ 0)$ can appear on a path from alginment vertex
  $(k_1, \ k_2)$ to the termination vertex if $j_1 - k_1 = j_2 - k_2$,
  the vertex $(j_1, \ j_2, \ 1)$ can appear if $j_1 - k_1 = j_2 - k_2 + 1$, 
  and the vertex $(j_1, \ j_2, \ 2)$ can appear if $j_1 - k_1 +1 = j_2 - k_2 $.
  If $j_1 = 1$ and/or $j_2 = 1$, then we associate the segmentation pair
  $(d_1 (j_1 ), \   d_2 (j_2 ))$ with the vertex  $(j_1, \ j_2, \ 0)$.
  Let $W(j_1, \ j_2, \ v)$ denote the weight of any incoming edge to
  $(j_1, \ j_2, \ 0), \; (j_1, \ j_2, \ 1),$ and $(j_1, \ j_2, \ 2); \; v$
  is a positive parameter that discourages the selection of the edge in an
  optimal path when $|d_1 (j_1 )| < v$ and/or $|d_2 (j_2 )| < v$,

  The outgoing edges from alignment vertex $(j_1, \ j_2)$ are to \\
  $\{(j_1 +k, \ j_2+k, \ 0): k \in \{1, \ 2, \ \dots , \ N- \max \{ j_1 , \
  j_2 \} \} \}$.

  The outgoing edges from vertex $(j_1, \ j_2, \ 0)$ are to
  \begin{itemize}
  \item the termination vertex with edge weight zero
    \item   $\{(j_1 +k, \ j_2+k, \ 0): k \in \{1, \ 2, \ \dots , \ N- \max \{ j_1 , \
      j_2 \} \} \}$ if $j_1 < N$ and $j_2 < N$
\item   $\{(j_1 +k+1, \ j_2+k, \ 1): k \in \{1, \ 2, \ \dots , \ N- \max \{ j_1 + 1 , \
  j_2 \} \} \}$ if $j_1 < N-1$ and $j_2 < N$
      \item   $\{(j_1 +k, \ j_2+k+1, \ 2): k \in \{1, \ 2, \ \dots , \ N- \max \{ j_1 , \
      j_2 +1 \} \} \}$ if $j_1 < N$ and $j_2 < N-1.$
    \end{itemize}

  The outgoing edges from vertex $(j_1, \ j_2, \ 1)$ are to
  \begin{itemize}
  \item the termination vertex with edge weight zero
    \item   $\{(j_1 +k, \ j_2+k, \ 1): k \in \{1, \ 2, \ \dots , \ N- \max \{ j_1 , \
      j_2 \} \} \}$ if $j_1 < N$ and $j_2 < N$
 \item   $\{(j_1 +k, \ j_2+k+1, \ 0): k \in \{1, \ 2, \ \dots , \ N- \max \{ j_1 , \
      j_2 +1 \} \} \}$ if $j_1 < N$ and $j_2 < N-1.$
  \end{itemize}

    The outgoing edges from vertex $(j_1, \ j_2, \ 2)$ are to
  \begin{itemize}
  \item the termination vertex with edge weight zero
    \item   $\{(j_1 +k, \ j_2+k, \ 2): k \in \{1, \ 2, \ \dots , \ N- \max \{ j_1 , \
      j_2 \} \} \}$ if $j_1 < N$ and $j_2 < N$
\item   $\{(j_1 +k+1, \ j_2+k, \ 0): k \in \{1, \ 2, \ \dots , \ N- \max \{ j_1 + 1 , \
  j_2 \} \} \}$ if $j_1 < N-1$ and $j_2 < N$
  \end{itemize}

  Define the edge weight function
          \begin{displaymath}
  W_1 (j_1, \ j_2, \ v)          
            = \left\{ \begin{array}{ll}
              1,        & d_1 [j_1 ] d_2 [j_2 ] > 0, \; |d_i [j_i ]| \geq v \;
              \mbox{for} \; i\in \{1, \ 2\}   \\
  0 ,     & \mbox{otherwise,}
\end{array}
\right.
    \end{displaymath}

          We have the following result.

          \noindent {\bf Theorem 5:} Suppose $d_i, \ i \in \{1, \ 2\}$
          satisfy the conditions of Theorem~4.  Then the longest path scheme
          with edge weights $W_1 (j_1, \ j_2, \ v)$ identifies the location
          of the first sample following each discontinuity point of $g(t)$
          within $\gamma_i$ and $y_i , \ i \in \{1, \ 2\}$.
          Moreover, it will not propose false segmentation points.

          \noindent {\bf Proof:} By the conditions on $d_1$ and $d_2$, the
          $N-m-1$ zero components of $\delta_i$ correspond
          to the $N-m-1$ components of $d_i$ with magnitude less than $v$,
          and $d_i$ preserves the signs of the $m+1$ non-zero components of
          $\delta_i$.  Define the indicator vectors
          $\iota_i = (\iota_i [1], \ \dots , \ \iota_i [N]), \ i \in \{1, \ 2\}$,
          by
          \begin{displaymath}
\iota_i [j]
            = \left\{ \begin{array}{ll}
              \delta_i [j] / | \delta_i [j] |,
              & \mbox{if} \; \delta_i [j] \neq 0 \\
  0 ,     & \mbox{if} \; \delta_i [j] = 0.
\end{array}
\right.
          \end{displaymath}
          Then by Theorem 4, $\iota_i , \; i \in \{1, \ 2\},$ can occur at
          most once on any path from the starting vertex to the termination
          vertex.  By Cauchy's inequality, the longest path can have weight
          at most $m+1$.  If the only edges of weight zero used in a longest
          path are those with the starting or the termination vertices as an
          endpoint, then there is a unique longest path with weight $m+1$ that
          proceeds from the starting vertex to the alignment vertex for
          $\gamma_1$ and $\gamma_2$ to a vertex corresponding to the first
          non-zero components of $\delta_1$ and $\delta_2$
          to a vertex corresponding to the second
          non-zero components of $\delta_1$ and $\delta_2$ and so on until
          traveling from the vertex corresponding to the last
          non-zero components of $\delta_1$ and $\delta_2$ to the termination
          vertex. $\Box$

          For our motivating example, if $0 \leq x < 0.5$ and $v=1$, then the
          optimal path corresponding to edge weight function
          $  W_1 (j_1, \ j_2, \ v)$ proceeds from the starting vertex to vertex
          $(0, \ 0)$ to vertex $(2, \ 2, \ 0)$ to vertex $(4, \ 3, \ 1)$
              to vertex $(5, \ 5, \ 0)$ to vertex $(7, \ 6, \ 1)$
                  to vertex $(8, \ 7, \ 1)$ to the termination vertex,
                    and it provides the correct alignment and segmentation
                    positions for $d_1$ and $d_2$.

                    Next suppose that $x=0.5$.  If we use $  W_1 (j_1, \ j_2, \ v)$ with
                    $1<v \leq 2$, then only one component of $d_2$ can
                    contribute to a positive edge weight, and the method fails.
                    If $0.5<v \leq 1$, then the algorithm correctly identifies
                    the alignment between the sequences, but there are two
                    candidates for the longest path which both incorrectly
                    estimate five constant regions for the support of $g(t)$
                    instead of four.  The two candidate longest paths are
                    \begin{eqnarray*}
& &                      \mbox{the starting vertex} \; \longrightarrow \; (0, \ 0) \; \longrightarrow \; (2, \ 2, \ 0) \; \longrightarrow \; (3, \ 3, \ 0) \\
                      & &                      \longrightarrow \; (4, \ 4, \ 0)
                      \; \longrightarrow \; (5, \ 5, \ 0) \; \longrightarrow \; (6, \ 6, \ 0) \\
& & \longrightarrow \; (8, \ 7, \ 1) \; \longrightarrow \;
\mbox{the termination vertex}
                    \end{eqnarray*}
                    and
                                        \begin{eqnarray*}
& &                      \mbox{the starting vertex} \; \longrightarrow \; (0, \ 0) \; \longrightarrow \; (2, \ 2, \ 0) \; \longrightarrow \; (3, \ 3, \ 0) \\
                      & &                      \longrightarrow \; (4, \ 4, \ 0)
                      \; \longrightarrow \; (5, \ 5, \ 0) \; \longrightarrow \; (7, \ 6, \ 1) \\
& & \longrightarrow \; (8, \ 7, \ 1) \; \longrightarrow \;
\mbox{the termination vertex}
                                        \end{eqnarray*}
                                        Between them, they yield three candidate estimates for a translation of
                                        $g(t)$, the best of which comes from applying the second longest path with $l=3$.  
          The energy of the difference of $\hat{g}^{(2)} (t)$ at $x=0$
        and  $\hat{g}^{(3)} (t)$ for the best solution when $x=0.5$
        is $\frac{41T}{144}$.

        This example illustrates two more issues.  First, when the noise
        level is sufficiently high, the longest path algorithm produces an
        estimate of $g(t)$ with too many regions.  Although this work is
        motivated by quantization the additive white Gaussian noise model
        is widely studied.  If the initial noise terms $e_1$  and $e_2$
        are independent and identically distributed zero-mean Gaussian
        random variables with variance $\sigma^2$, and if $j_1 \geq 2, \
        j_2 \geq 2, \ \delta_1 [j_1 ] = a$ and $\delta_2 [j_2 ] = b$, then the
        probability that the corresponding weight in the graph is positive
        can be shown to be
        \begin{displaymath}
          Q \left( \frac{v-a}{\sigma \sqrt{2}} \right)
         Q \left( \frac{v-b}{\sigma \sqrt{2}} \right)
          + Q \left( \frac{v+a}{\sigma \sqrt{2}} \right)
         Q \left( \frac{v+b}{\sigma \sqrt{2}} \right),
        \end{displaymath}
        where $Q(z) = \int_z^{\infty} \frac{e^{-t^2 / 2}}{\sqrt{2 \pi}} dt.$
        This is always positive, so the algorithm may propose false
        segmentation point that would tend to increase the errors in the
        estimate of $g(t)$.

        Our motivating example at $x=0.5$ also highlights the problem of two
        longest paths, one of which uses the vertex $(7, \ 6, \ 1)$
        corresponding to $d_1[7] = -2$ and the other which uses the vertex
        $(6, \ 6, \ 0)$ associated with $d_1[6] = -1$.
        If the noise does not dominate the uncorrupted signal, then we tend to infer
        that the components of $d_i, \ i \in \{1, \ 2\}$, with larger
        magnitudes are likely to correspond to segmentation points,
        and this is not fully captured by the edge weight function
        $  W_1 (j_1, \ j_2, \ v)$.  Therefore, it is natural to consider a
        second edge weight function
            \begin{displaymath}
  W_2 (j_1, \ j_2, \ v)          
            = \left\{ \begin{array}{ll}
d_1 [j_1 ] d_2 [j_2 ] ,  & d_1 [j_1 ] d_2 [j_2 ] > 0, \; |d_i [j_i ]| \geq v \;
              \mbox{for} \; i\in \{1, \ 2\}   \\
  0 ,     & \mbox{otherwise.}
\end{array}
\right.
            \end{displaymath}

            For the preceding additive white Gaussian noise model, it is
            possible to show that when 
$j_1 \geq 2, \
        j_2 \geq 2, \ \delta_1 [j_1 ] = a$ and $\delta_2 [j_2 ] = b$, we have \\$         E[   W_2 (j_1, \ j_2, \ v) ]$
        \begin{eqnarray*}
  & = &
          \left( aQ \left( \frac{v-a}{\sigma \sqrt{2}}\right)
+ \frac{\sigma}{\sqrt{\pi}} e^{- \frac{(v-a)^2}{4 \sigma^2} } \right)
                    \left( bQ \left( \frac{v-b}{\sigma \sqrt{2}} \right)
+ \frac{\sigma}{\sqrt{\pi}} e^{- \frac{(v-b)^2}{4 \sigma^2} } \right) \\
          & &  +           \left( aQ \left( \frac{v+a}{\sigma \sqrt{2}} \right)
- \frac{\sigma}{\sqrt{\pi}} e^{- \frac{(v+a)^2}{4 \sigma^2} } \right)
                    \left( bQ \left( \frac{v+b}{\sigma \sqrt{2}} \right)
- \frac{\sigma}{\sqrt{\pi}} e^{- \frac{(v+b)^2}{4 \sigma^2} } \right) .
        \end{eqnarray*}
        When $a$ and $b$ either both exceed $v$ or are both less than $-v$,
        the preceding expression approaches $ab$ as $\sigma$ decreases to zero.

        Although $W_2 (j_1, \ j_2, \ v)$ emphasizes components of
        $d_1$ and $d_2$ with large magnitudes within the optimization,
        it may or may not properly align such components in the presence of
        noise.  Therefore, we propose a simple variation of Cauchy's inequality
        and a third edge weight function which we conjecture would be helpful
        with additive white Gaussian noise.

\noindent {\bf Proposition 6:} For sequences
of real numbers $\{ a_i \}$  and $\{ b_i \}$ with $a_i b_i \geq 0$ for all $i$,
\begin{equation}
\sum_i \min \{{a_i}^2 , \ {b_i}^2 \} \; \leq \; \sum_i a_i b_i \; \leq \; \sqrt{ \sum_i {a_i}^2 \sum_i {b_i}^2 }
\end{equation}
with equality if and only if $a_i = b_i$ for all $i$.

The corresponding edge weight function is
            \begin{displaymath}
  W_3 (j_1, \ j_2, \ v)          
            = \left\{ \begin{array}{ll}
              \min \{ (d_1 [j_1 ])^2, ( d_2 [j_2 ])^2 \} ,  & d_1 [j_1 ] d_2 [j_2 ] > 0, \\
              & |d_i [j_i ]| \geq v \;
              \mbox{for} \; i\in \{1, \ 2\}   \\
  0 ,     & \mbox{otherwise.}
\end{array}
\right.
            \end{displaymath}

            It is possible to make a crude probabilistic comparison between \\
            $  W_2 (j_1, \ j_2, \ v)$
            and          $  W_3 (j_1, \ j_2, \ v)$ when the initial noise terms
            $e_1$ and $e_2$ are independent and identically distributed zero-mean uniformly distributed continuous random variables.
            
\section{Conclusions} 
Counterexamples are part of the tradition of mathematical discovery
(see, e.g., \cite{analysis}-\cite{probability}).  Selecting the maximum
from cross-correlation template matching is a well-known alignment strategy
for discrete image registration, but it is not always the best alignment
technique for noisy signals.  Moreover, it does not address the need for
segmentation in sequences of noisy samples from quantized signals.
Motivated by an example, we prove that schemes based on difference sequences,
thresholding, and in some cases dynamic programming offer optimal alignment
and segmentation for sequences of noisy samples from a one-dimensional spatially
limited piecewise constant function under some conditions on the noise.
We also consider some of the challenges in a more general case.

\end{document}